\documentclass[final,3p,times,twocolumn]{elsarticle}

\usepackage{lineno}

\usepackage{tabularx}
\usepackage{todonotes}
\presetkeys{todonotes}{inline}{}  

\usepackage{url}
\usepackage{siunitx}

\usepackage[inkscapelatex=false]{svg}
\usepackage{amsmath}
\usepackage{float}
\usepackage{multirow}
\usepackage{comment}

\usepackage[labelfont=bf, labelsep=period]{caption}
\usepackage{subcaption}

\usepackage{amssymb}

\usepackage{xcolor}
\usepackage{booktabs}
\definecolor{light-gray}{gray}{0.95}
\newcommand{\code}[1]{\colorbox{light-gray}{\small\texttt{#1}}}

\usepackage{hyperref}
\hypersetup{colorlinks, allcolors=black}

\usepackage[utf8]{inputenc}
\journal{}

\begin{document}

\begin{frontmatter}

\title{Forecasting COVID-19 spreading trough an ensemble of classical and machine learning models: Spain's case study}





\author[IFCA]{Ignacio Heredia Cacha}
\ead{iheredia@ifca.unican.es}
\author[IFCA]{Judith Sáinz-Pardo Díaz}
\ead{sainzpardo@ifca.unican.es}
\author[IFCA]{María Castrillo Melguizo}
\ead{castrillo@ifca.unican.es}
\author[IFCA]{Álvaro López García\corref{correspondingauthor}}
\ead{aloga@ifca.unican.es}

\address[IFCA]{Instituto de Física de Cantabria (IFCA), CSIC-UC \\
Avda. los Castros s/n. 39005 - Santander (Spain)
}

\cortext[correspondingauthor]{Corresponding author}

\begin{abstract}

In this work we evaluate the applicability of an ensemble of population models and machine learning models to predict the near future evolution of the COVID-19 pandemic, with a particular use case in Spain. We rely solely in open and public datasets, fusing incidence, vaccination, human mobility and weather data to feed our machine learning models (Random Forest, Gradient Boosting, k-Nearest Neighbours and Kernel Ridge Regression). We use the incidence data to adjust classic population models (Gompertz, Logistic, Richards, Bertalanffy) in order to be able to better capture the trend of the data. We then ensemble these two families of models in order to obtain a more robust and accurate prediction. Furthermore, we have observed an improvement in the predictions obtained with machine learning models as we add new features (vaccines, mobility, climatic conditions), analyzing the importance of each of them using Shapley Additive Explanation values. As in any other modelling work, data and predictions quality have several limitations and therefore they must be seen from a critical standpoint, as we discuss in the text. Our work concludes that the ensemble use of these models improves the individual predictions (using only machine learning models or only population models) and can be applied, with caution, in cases when compartmental models cannot be utilized due to the lack of relevant data.

\end{abstract}


\end{frontmatter}


\section{Introduction}
\label{sec:introduction}

After the surge of cases of the new Coronavirus Disease 2019 (COVID-19), caused by the SARS-COV-2 virus, in every region in Spain by the second week of March 2020 several measures were imposed to slow down the spread of the disease. Over the time these measures have included hard lock-downs, restrictions on people's mobility, limiting the number of people public places and the usage of protection gear (masks or gloves), among others.

In addition, the application of those measures has not been consistent between countries nor between Spain's regions. This makes it hard to reliably assess the impact of the individual restrictions in stopping the spreading \cite{aloi2020effects,mazzoli2020effects}. Human mobility and its direct impact on the spread of infectious diseases (including COVID-19) has been profusely studied, and restricting or limiting the mobility from infected areas is one of the first measures being adopted by authorities in order to prevent an epidemic spread, with different results \cite{mazzoli2020effects,10.1371/journal.pcbi.1009326,doi:10.1126/science.abc5096,meloni2011modeling,ferguson2005strategies,JRC121300,ponce2021covid}. Moreover, with the aim of trying to infer if there is any relationship between weather and the increase or decrease of COVID-19 infections, we have made an analysis of weather data. The motivation for this is that it is known that other respiratory viruses survive less in humid climates and with low temperatures \cite{clima_covid}. In addition, some studies already attempt to relate the influence of climate on COVID-19 cases, for example \cite{ROSARIO2020113587}, where it is concluded that climatic factors play an important role in the pandemic, and \cite{sharma2020correlation}, where it is also concluded that climate is a relevant factor in determining the incidence rate of COVID-19 pandemic cases (in the first citation this is concluded for a tropical country and in the second one for the case of India). For this, we have used the daily data per meteorological station provided by the Spanish Meteorological Agency ---in Spanish \textit{Agencia Española de Meteorología (AEMET)}---

Our approach aims to predict the spread of COVID-19 combining both machine learning and classical population models, using exclusively publicly available data (incidence, mobility, vaccination and weather). Having a reliable forecast will enable us to assess the influence of these factors on the spreading rate, thus allowing decision makers to design more effective policies.

The motivation for using these two types of models stems from the fact that, in the tests we have performed, we have noticed that while machine learning models in the vast majority of cases overestimate the number of daily cases, population models generally seem to predict fewer cases than actual ones. To make the most both model families, we aggregate their predictions using ensemble learning. In ensemble learning, we optimally combine all the individual predictions to generate a meta-prediction. The ensemble usually outperforms any of its individual model members \cite{Opitz1999,Rokach2009}.

Our contribution is twofold. First we will combine both classical and ML predictions and study their optimal temporal range of applicability. Second, as classical models we will use less explored population growth models. Contrary to compartmental epidemiological models, these models can be used even when the data of recovered population are not available. This is a crucial advantage because recovered patient data are usually hard to collect, and in fact not available anymore for Spain since 17 May 2020 (see dataset in \cite{DONG2020533}).  It should be noted nevertheless that some regions do provide these data on recoveries and/or active cases, and there are some very successful works in the development of this type of compartmental models \cite{AREA2021559}.

The paper is structured as follows: Section~\ref{sec:related} contains the related work relevant to this publication; Section~\ref{sec:data} outlines the datasets considered for our work, as well as the pre-processing that we have performed to them; in Section~\ref{sec:methodology} we present the ensemble of models being used to predict the evolution of the epidemic spread in Spain; Section~\ref{sec:results} describes our main findings and results; and Section~\ref{sec:conclusions} contains our conclusions and future work.

\section{Related work}
\label{sec:related}

Much effort has been done to try predict the COVID-19 spreading, and therefore be able to design better and more reliable control measures \cite{Rdulescu2020}.
Many of the most solid work comes from classical compartmental epidemiological models like SEIR, where population is divided in different compartments (\textbf{S}usceptible, \textbf{E}xposed, \textbf{I}nfected, \textbf{R}ecovered). Many SEIR models have been extended to account for additional factors like confinements \cite{Lpez2021}, population migrations \cite{CHEN2020252}, types of social interactions \cite{Chung2021} or the survival of the pathogen in the environment \cite{Mwalili2020}. In particular, \cite{AREA2021559} predicts required beds at Intensive Care Units by adding 4 additional compartments to those of the SEIR model: Fatality cases, Asymptomatics, Hospitalized and Super-spreaders.

In this study, instead of compartmental models, we have chosen to use population models, for which we only need to have the data on the daily cases. Several papers already include the use of this type of models for the COVID-19 case studies, such as \cite{medina2020covid}, where the use of Gompertz curves and logistic regression is proposed, or \cite{brahma2021mathematical}, where the Von Bertalanffy growth function (VBGF) is used to forecast the trend of COVID-19 outbreak. Additionally, \cite{conde2021comparison} compares the use of artificial neural networks and the Gompertz model to predict the dynamics of COVID-19 deaths in Mexico. However, our approach will try not to compare the performance of both kind of models (data based and population models), but instead combine them to try to obtain more accurate and robust predictions.

In the recent years machine learning (ML) has emerged as a strong contender to classical mechanistic models. In the context of the spread of COVID-19 during the early phases of the outbreak, the focus was on trying to predict the evolution of the time series of pandemic numbers \cite{BOCCALETTI2020110278,9099302},
with disparate prediction quality and uncertainties. ML has been used both as a standalone model \cite{le2020neural} or as a top layer over classical epidemiological models \cite{Ark2021}. ML models have been used to exploit different big data sources \cite{CHEW2021107417,electronics10243125} or incorporating heterogeneous features \cite{9179729}. Also, several general evaluations of the applicability of these models exist \cite{CHAKRABORTI2021142723,KUO2021144151,ZEROUAL2020110121,VERMA2022116611}.

Regarding the model ensemble, work has been taking place both in the USA \cite{covid19_usa_ensemble} and EU \cite{covid19_eu_ensemble} to consolidate all these different models by deploying portals that ensemble the predictions. ML techniques have also been used to help improving classical epidemiological models \cite{s21020540}.

Despite everyone best efforts, sensible work has carefully warned against the possibility of meaningfully predicting the evolution for temporal horizons over a week \cite{Castro2020}, just as is the case for the weather. For this reason, we will do our best all over this paper to point at the limitations of our data (Section \ref{subsec:data_limitations}) and models so that we do not add more fuel to the hype wagon.

\section{Data}
\label{sec:data}

In the spirit of Open Science, all this work exclusively relies on open-access public data. We hope that, on the one hand this will contribute to the rigorous assessment of the models before they can be adopted by policy makers, whilst on the other hand it will encourage the release of comprehensive and quality open datasets by public administrations, not limited to the COVID-19 pandemic data.

Our dataset is composed of COVID-19 cases data, COVID-19 vaccination data, human population mobility data and weather observations, and is constructed as explained in what follows.

The spatial basic units of the present work are the whole country (Spain), and the autonomous community (Spain is composed of 17 autonomous communities and 2 autonomous cities). Due to their particular geographical situation and demographics, the pandemic outbreak in the two autonomous cities of Ceuta and Melilla had a different behaviour and they have not been analyzed individually in this study as has been done with the 17 autonomous communities, as they are considered as outliers. However, we have considered the daily cases reported by these autonomous cities in the total number of daily cases in Spain. Furthermore, in the case of mobility and temperature, these data are different if the analysis is carried out for the whole of Spain, or if it is done by Autonomous Region.

The dataset time range goes from the January 1, 2021 to December 31, 2021. For consistency, we do not include data before that date because vaccination in Spain started on December 27, 2020. Also, note that after November 2021, the daily cases exploded due to Omicron variant (cf. Figure~\ref{fig:cases_SP_CB}), so the forecasts will therefore be presumably worse in that month.

In the case of the machine learning models, these data will be divided into training, validation and test. Specifically, the days to be predicted in test will be, from October 2, 2021 (so the date on which the prediction would be made is October 1), until December 31. The 30 days prior to these dates will correspond to the validation set, and the rest to the training set. Note that forecasts will be made for 14 days. In the case of the population models, we consider the same test set, and as training the 30 days prior to the 14 days to be predicted (more details in Section~\ref{sec:pop_models}). 

\subsection{Daily COVID-19 cases data}

Concerning the data on daily cases confirmed by COVID-19, we use the data collected by the Carlos III Health Institute ---in Spanish \textit{Instituto de Salud Carlos III (ISCIII)}--- which is an Spanish autonomous public organization currently dependent on the Ministry of Science and Innovation ---in Spanish \textit{Ministerio de Ciencia e Innovación (MICINN)}---. The data source is available at \cite{data_isciii}.



The dataset separates new cases by the test technique used to detect them (PCR, antibody, antigen, unknown) and the autonomous community of residence. For this study, we use the total number new cases across all techniques. Therefore, the final objective will be to predict the number of daily cases per day for Spain as a whole and for each autonomous community.

\begin{figure}[H]
    \centering
    \vspace*{-1cm}
    \includesvg[width=\linewidth]{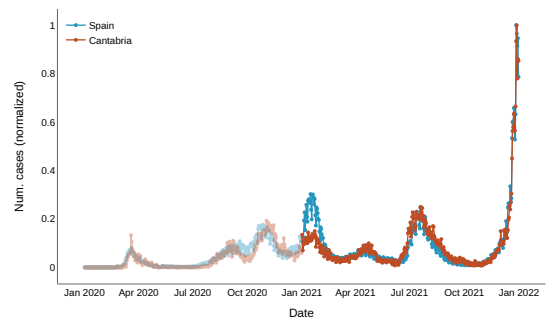}
    \caption{Daily COVID-19 confirmed cases (normalized) in Spain and in Cantabria autonomous community. Transparency is added to data outside our considered time range (data before 2021).}
    \label{fig:cases_SP_CB}
\end{figure}

Figure~\ref{fig:cases_SP_CB} shows the evolution of daily COVID-19 cases (normalized) throughout 2021 for Spain, and for the autonomous community of Cantabria. It should be noted that the evolution of the trend for Cantabria is analogous to that of the country as a whole.

\begin{figure}[H]
    \centering
    \vspace*{-1cm}
    \includesvg[width=\linewidth]{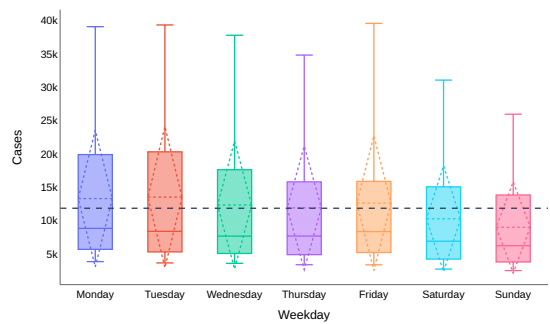}
    \caption{Statistics on the number of cases depending on the day of the week (ML train set). The dotted black line shows the mean of the daily cases in the study period, and in each boxplot the mean and standard deviation are also shown as dashed lines.}
    \label{fig:boxplots_weekday}
\end{figure}

Figure~\ref{fig:boxplots_weekday} shows the number of diagnosed cases according to the day of the week. We see that the data has (expected) weekly patterns: less cases are diagnosed on weekends. Although at first glance the patterns seem noticeable, after performing some test with the models which will be exposed later we finally decided to not include this variable, the day of the week, as input for the model.

\subsection{Vaccination data}
\label{sec:vaccination}

Vaccination against COVID-19 is essential to protect the most vulnerable groups, since its main objective is to reduce the severity and mortality of the disease.

The vaccination process in Spain began on December 27, 2020, prioritizing its inoculation to people living in elderly residences and other dependency centers, health personnel and first-line healthcare partners, and people with a high degree of dependency not institutionalized. The vaccination strategy continued with the most vulnerable people following an age criterion, in a descending order. By June 2021, the vaccine was widely available, and the process continued again in descending order of age, reaching those over 12 years of age. Thus, by October 14, 87.9$\%$ of the target population (i.e. those over 12 years old) had received the full vaccination schedule \cite{vac_spain}.

As of December 15, 2021, 4 vaccines were authorized for administration by the European Medicines Agency (EMA) \cite{vac_spain} (cf. Table~\ref{tab:vaccines_summary}).

\begin{table}[H]
    \centering
    \resizebox{\linewidth}{!} {
    \begin{tabular}{cccc}
    \toprule
         Abbrv. & Company & Vaccine type & Dosage\\
         \midrule
         AZ & AstraZeneca & Adenovirus vector & 2 doses\\
         COM & Pfizer/BioNTech & ARNm & 2 doses\\
         JANSS & Janssen & Adenovirus vector & 1 dose\\
         MOD & Moderna & ARNm & 2 doses\\
    \bottomrule
    \end{tabular}}
    \caption{Vaccines authorized by the EMA.}
    \label{tab:vaccines_summary}
\end{table}

The data from the Ministry of Health of the Government of Spain on the vaccination strategy consists of reports on the evolution of the strategy, i.e., no daily or weekly data on the doses administered are publicly available. Therefore, in this study we will use the European COVID-19 vaccination data collected by the European Centre for Disease Prevention and Control. This dataset contains the doses administered per week in each country, by vaccine type and age group. In addition, a distinction is made between whether the vaccine corresponds to a first or second dose in each case. The data source is available in \cite{data_vac}.

\begin{figure*}[ht!]
    \centering
    \vspace*{-0.75cm}
    \includesvg[width=\textwidth]{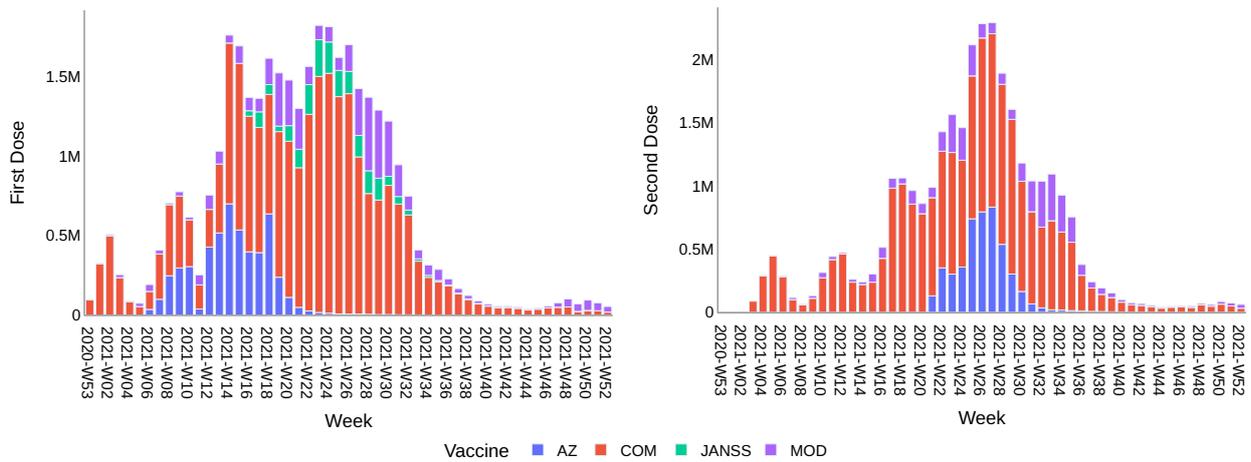}
    \caption{First and second doses of the COVID-19 vaccine given in Spain by week and type of vaccine.}
    \label{fig:Vaccines_1st2nd_doses}
\end{figure*}

In Figure~\ref{fig:Vaccines_1st2nd_doses} we show the weekly evolution of the vaccination strategy considering the type of vaccine, and the first and second doses (without distinguishing by age groups).

As the number of doses administered is given at weekly level (i.e. doses administered each week), we are interested in extrapolating these data to the daily level.

\begin{figure}[H]
    \centering
    \vspace*{-1cm}
    \includesvg[width=\linewidth]{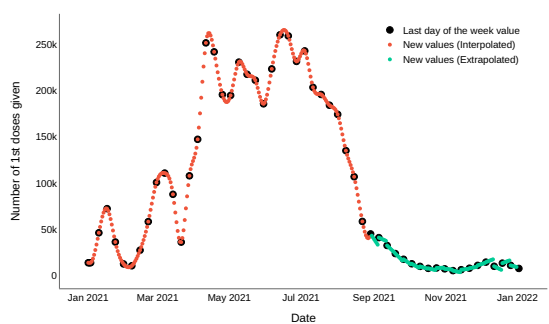}
    \caption{Interpolated and extrapolated values for each day of 2021 for the first dose of the vaccine.}
    \label{fig:Vaccines_interp_extrap_train_spl3}
\end{figure}

As the value of the total weekly doses is not known until the last day of each week, we associate to each Sunday the total value of doses administered that week divided by 7. Now, we must assign values for the intermediate days. Note that, in order to predict the cases of day $n$, the vaccination, mobility and climate data on day $n-14$ will be used (the motivation for this is explained in Section~\ref{sec:ml-models}, and it is also shown in Table~\ref{tab:input_14days}). Then, in order not to use future data in our test set (we don´t know the data from the last available day to $n$), we cannot interpolate those values for that part of the data, so the process to follow has been the following: we \textit{interpolate} using cubic splines with the known data until August 29, 2021 (the training set covered up to September 1, 2021), and from the last known data, we \textit{extrapolate} linearly until the end of that week (when a new data will be available). That is, if we consider as known days the last day of each week, every time we reach a new known data, we continue the linear extrapolation. The result obtained for the data of the first dose is shown in Figure~\ref{fig:Vaccines_interp_extrap_train_spl3}, where it can be seen which values were known because it was the last day of the week, which have been interpolated and which have been extrapolated.

Thus, through a process of interpolation (for the train set), and extrapolation for validation and test sets, we have associated to each day of 2021 a value for the vaccination data of the first and second doses of COVID-19 vaccine. Figure~\ref{fig:Vaccines_interp_extrap_train_spl3} shows the result corresponding to the first dose, but the process followed for the second dose data has been analogous.

\subsection{Mobility data}
\label{sec:mobility}

In order to asses human mobility we used the data provided by the Spanish National Statistics Institute ---in Spanish \textit{Instituto Nacional de Estadística (INE)}---.
The data source is available at \cite{data_ine}.

Since 2019 the INE has conducted a mobility measurement project from mobile telephony. In 2020, during the period corresponding to the state of alarm, due to the impact of mobility in the COVID-19 pandemic in Spain, this project provided daily information on movements between the 3214 mobility areas that were designed for the project. For this period, from March 16 to June 20, the telephone operators provided daily mobility data.
Subsequently, due to the continuous waves of the pandemic and the influence of mobility on its evolution, the study continued, but now with the publication of mobility data every week, relative to two specific days of the previous week (Wednesday and Sunday). Information on the study is available at \cite{data_ine}.

Regarding the data collected in this project, we are interested in knowing the flux between different population areas, for which we will have areas of residence and areas of destination.

Some important aspects of the data provided by this study are summarized below:
\begin{itemize}
    \item Mobile location data are obtained from the three major mobile operators in the country (Orange, Telefónica and Vodafone).
    \item The area of residence of each mobile phone is considered to be the area where it is located for the longest time between 22:00 hours on the previous day and 06:00 hours during the observed day.
    \item To determine the area of destination, all areas (including the residence one) in which the terminal is located during the hours of 10:00 to 16:00 of the observed day should be taken. If there is more than one area, the one where the terminal is located the longest, other than the area of residence, will be taken.
    \item To preserve user privacy, whenever the number of observations is less than 15 in an area, for a given operator, the result will be censored at source. Origin-destination mobility data is then only provided for areas in which at least one of the three operators passes this threshold.
    \item As in most of the study data are only available for two days of each week, a forward fill is performed when data was not available (i.e. propagating the known values as will be explained hereinafter).
\end{itemize}

\begin{figure}[H]
    \centering
    \includegraphics[width=\linewidth]{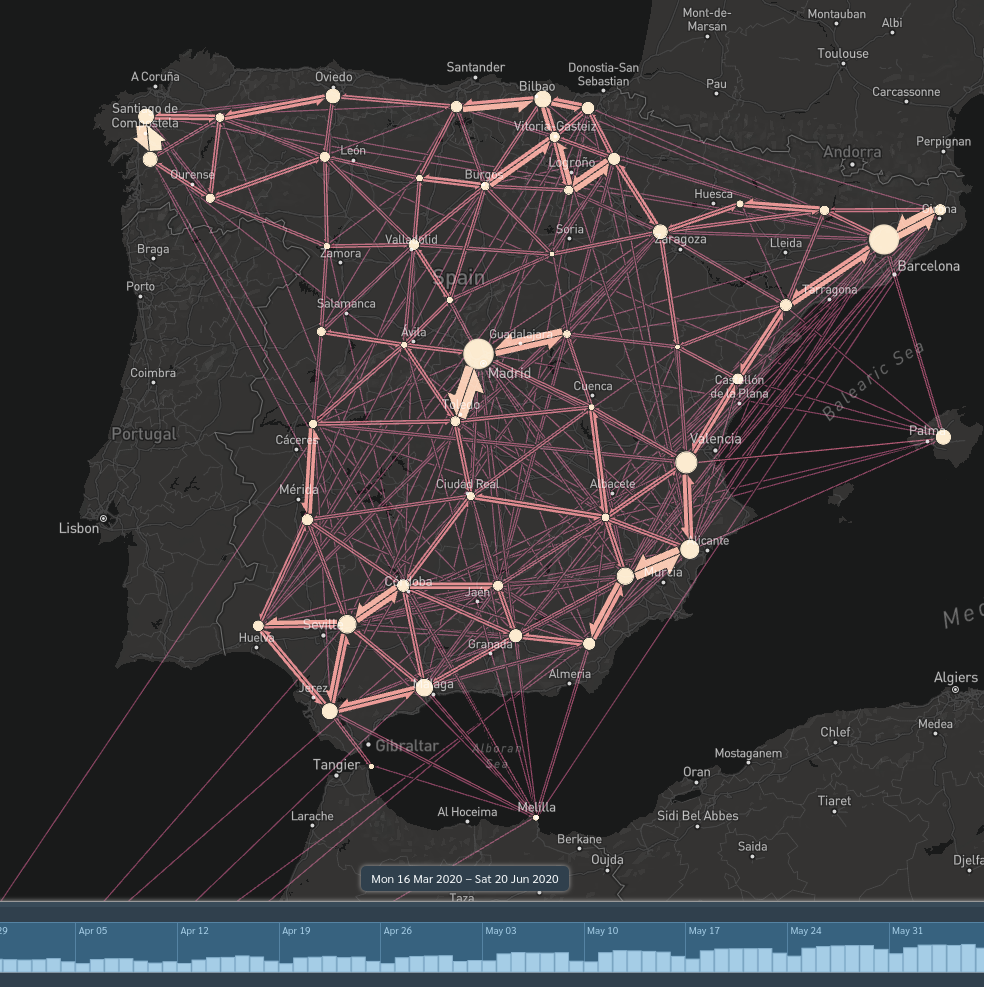}
    \caption{Mobility fluxes in Spain. Arrow size shows inter-province fluxes and dot size shows intra-province fluxes.}
    \label{fig:flux_spain}
\end{figure}

Specifically, the the origin-destination flows shown in Table 1.4 of the INE mobility data have been considered. Figure~\ref{fig:flux_spain} shows a visual representation of those fluxes.

\begin{figure}[H]
    \centering
    \vspace*{-1cm}
    \includesvg[width=\linewidth]{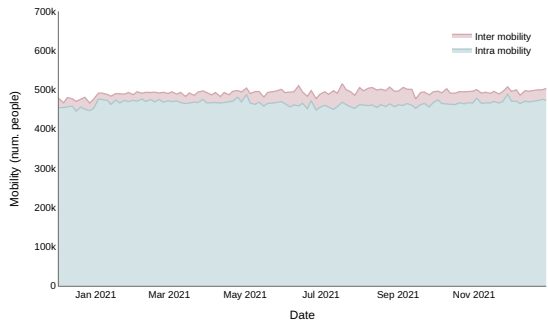}
    \caption{Mobility fluxes in Cantabria, separating the contributions of the two components: intra-mobility (people that move inside Cantabria) and inter-mobility (people that arrive to Cantabria).}
    \label{fig:flux_CB_area}
\end{figure}

We then have to assign, from all these fluxes, a mobility value to each Autonomous Community for each day. Be $X_i$ each of the $N$ autonomous communities considered in the study, $i \in \{1,\hdots,N\}$. The mobility flux assigned to an Autonomous Community $X_{i}$ on a given day $t$ ($F_{X_{i}}^{t}$) will be the sum of all the incoming fluxes from the remaining $N-1$ communities (inter-mobility), that is $f_{X_{j} \rightarrow X_{i}}^{t}$ $\forall j \in \{1,\hdots,N\}$, $j \neq i$, together with the internal flux  $f_{X_{i} \rightarrow X_{i}}^{t}$ inside that community (intra-mobility):
\begin{equation}
    F_{X_{i}}^{t} = \sum_{j=1}^{N} f_{X_{j} \rightarrow X_{i}}^{t}
\end{equation}
When studying the whole country, Spain, the mobility is just the sum of the flux of all the Autonomous Communities. Figure~\ref{fig:flux_CB_area} shows the temporal evolution of mobility for Cantabria, separating the intra-mobility and inter-mobility components.

\begin{figure}[H]
    \centering
    \includegraphics[width=\linewidth]{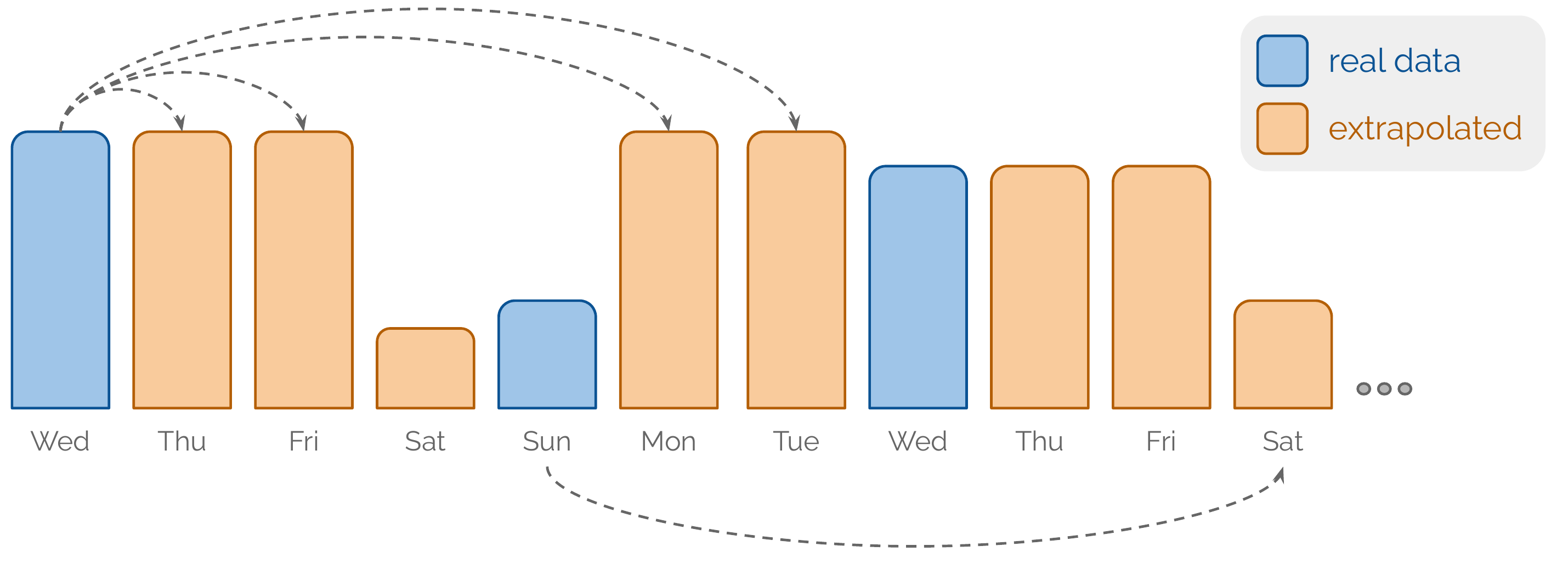}
    \caption{Mobility data processing.}
    \label{fig:mobility-processing}
\end{figure}

As real mobility data are only published for Wednesdays and Sundays, we have taken the following approach to assign daily mobility values to the remaining days. For each week, we assign Monday/Tuesday the values of previous Wednesday, Thursday/Friday the values of current Wednesday, and Saturday the value of previous Sunday. The process is shown in Figure~\ref{fig:mobility-processing}.

This approach is based in two key observations: (1) mobility has strong weekly pattern (high on weekdays, low on weekends); (2) We cannot directly assign Wednesday value for all weekdays in the week because that would create an information leak (i.e. on Monday one cannot already know Wednesday mobility); same argument goes for weekends. Avoiding this information leak is especially important in the test dataset, hence this approach.

\subsection{Weather conditions data}
\label{sec:weather}

Daily weather data records for Spain, since 2013, are publicly available \cite{bd_clima}.

As discussed in the introduction, we are mainly interested in temperature data, as there is evidence suggesting they could be linked to the infection rate of COVID-19. On the other hand, we were interested in using humidity data for the same reason. As humidity is not an available variable in the records, we use precipitation instead of it.
For assigning a daily temperature/precipitation value to each Autonomous Community we simply average the mean daily values of all stations located in that Autonomous Community. In the case of Spain, we take the average of all stations.

As we are mainly interested in seeing if large scale weather trends (mainly seasonal) have and influence of spreading, we have performed a 7-day rolling average of these values (both temperature and precipitations). This will also help reducing the noise in the input data for the models.

\subsection{Data limitations} \label{subsec:data_limitations}

Most of the data limitations are of course not exclusive to this paper. But we want nonetheless gather them all together so the reader can have a clearer picture of the confidence level on the results here found. Here are some of the limitations we came around while developing this work:

\begin{itemize}
    \item Incidence data in not always a good proxy of infected people because it relies on the number of diagnostic tests performed. This led to an underestimation of infected people especially at the beginning of the pandemic due to test not being widely available. Not performing tests on the whole population, just on symptomatic people, also leads to an underestimation of infected people. Holidays might also modify testing patterns.

    \item Incidence prediction can be reliable usually up to two weeks, but further predictions will be influenced by future data not yet available when making the predictions. These data includes future control measures, future vaccination trends, future weather, etc.
    Therefore measuring the accuracy of the model for time ranges beyond that limit is not a good assessment of its quality, that is why all results in this work are limited to 14-day forecasts.


    \item Vaccination data are only available weekly and given at country level, so fine-grained differences in vaccination levels between regions are lost.

    \item Spain is a regional state, so each autonomous community is the ultimate responsible of healthcare institutions, therefore leading sometimes to methodological disparities between administrations when reporting cases.

    \item Infection data does not report the COVID-19 variant. Therefore models have a limited time-range applicability. Models trained at the beginning of the pandemic will hardly be able to predict the high-rate spreading of the Omicron variant \cite{COVID_variants}, as we will see in the results.

    \item Mobility data can be misleading, as it does not always equate to risk of infection, as certain activities may show more risk of infection than others, regardless of the level of mobility required for each of them. For example, in \cite{ijgi10060395} it is mentioned that markets and other shopping malls with frequent visitors were areas with high risk of infection (in the case of Wuhan, China), so, in general, mobility to these types of places may suppose a higher exposure to the disease. In addition, we only have the actual data on Wednesdays and Sundays, from which we have to infer the values we will consider the rest of the days.

    \item The weather value of a region has been taken as the average of all weather stations located inside that region. Despite being a good first approximation, this is obviously not optimal. Stations located near densely populated areas should have greater weight than those located near sparsely populated areas.
\end{itemize}

\section{Methods}
\label{sec:methodology}

In this work we have designed an ensemble of models to predict the evolution of the epidemic spread in Spain. Specifically, we ensemble the predictions of both machine learning and population models.

\definecolor{green}{RGB}{2, 88, 16}
\definecolor{red}{RGB}{162, 31, 34}
We purposely decided to use population models instead of the classical SEIR models (which are designed to model pandemics) because Spain no longer publishes the data of recovered patients. These daily recoveries (or the daily number of active cases) is crucial in order to estimate the recovery rate, and thus SEIR's basics compartments (\textbf{S}usceptible, \textbf{E}xposed, \textbf{I}nfected, \textbf{R}ecovered).
As can be seen in the following equation, the \textcolor{red}{missing data} cannot be inferred from  \textcolor{green}{available data}, so the data on the daily recovered are not available:
$$ \textcolor{green}{Confirmed} =
  \textcolor{red}{Active}
+ \textcolor{red}{Recovered}
+ \textcolor{green}{Deceased}$$

In this study we use a training set to train the ML models and fit the parameters of the population models. To ensemble the models, the predictions of each model for the test set are weighted according the root-mean-square error (RMSE) performance in the validation set.

\subsection{Models definition}

\subsubsection{Population Models}
\label{sec:pop_models}

Population models are mathematical models applied to the study of population dynamics. The classic application of this kind of models is to analyze and predict the growth of a population (i.e. the census of a certain place) \cite{meade1988modified}. However, there are numerous applications in other fields, from animal growth \cite{bertalanffy_fish}, tumor growth \cite{fernandez2021mathematical}, evolution of plant diseases \cite{gompertz_plants}, etc. In addition, several papers use this type of model to try to predict the future trend of COVID-19 cases, as exposed in Section~\ref{sec:related}.

Specifically in this study, we will use the following four models.

\begin{description}
    \item[Gompertz model]
        is a type of mathematical model which is described by a sigmoidal function, so that growth is given as slower at the beginning and end of the time period studied. It is used in numerous fields of biology, from modeling the growth of animals and plants to the growth of cancer cells \cite{tjorve2017use}.

        Be $p(t)$ the population at time t, then, the ordinary differential equation (ODE) which defines the model is given by:
        \begin{equation}
            \frac{\partial p}{\partial t} = ap(t) -bp(t)log(p(t))
        \end{equation}

        And its explicit solution:
        $$
        \boxed{p(t) = e^{\frac{a}{b}+c e^{-bt}}}
        $$
        \begin{description}
            \item[Optimized parameters]
                Once we have the explicit solution for the ODE of the model, we need to estimate the three parameters involved: $a$, $b$ and $c$. To do so, we follow the process described in \ref{appendix_gomp}. When we get an initial estimation for $a$, $b$ and $c$, these parameters are optimized using the explicit solution of the ODE and the known training data. Specifically in our study we have used the sum of squares of the error for this purpose.
            \item[Implementation] for the optimization of parameters from the initial estimation, \code{fmin} function from the \code{optimize} package of \code{scipy} library \cite{2020SciPy-NMeth} has been used.
        \end{description}

    \item[Logistic model]
        This model introduced by Verhulst in 1838 \cite{verhulst1838notice} establishes that the rate of population change is proportional to the current population $p$ and $K-p$, being $K$ the carrying capacity of the population. Thus, be $a$ the constant of proportionality, and $b =\frac{a}{K}$, the ODE that defines the model it is given by:
        \begin{equation}
            \frac{\partial p}{\partial t} = ap(t)-bp^{2}(t)
        \end{equation}
        And the explicit solution:
        $$
        \boxed{p(t) = \frac{1}{c e^{-at}+\frac{b}{a}}}
        $$

        Again it will be necessary to calculate some initial parameters (which will later be optimized as we explained in the case of the Gompertz model) $a$, $b$ and $c$.

        \begin{description}
            \item[Optimized parameters] $a$, $b$ and $c$, first estimated following an analogous process to that of the Gompertz model.

            \item[Implementation] for the optimization of the initial parameters \code{fmin} function from the \code{optimize} package of \code{scipy} library \cite{2020SciPy-NMeth} has been used.
        \end{description}

    \item[Richards model]
        it is a generalization of the logistic model or curve \cite{WANG201212}, introducing a new parameter, $s$, which allows greater flexibility in the modeling of the curve. It is defined by the following EDO:
        \begin{equation}
            \frac{\partial p}{\partial t} = \frac{a}{s}p(t)\left(1-\left(\frac{p(t)}{p_{\infty}}\right)^{s}\right)
        \end{equation}

        And the explicit solution:
        $$
        \boxed{p(t) = \frac{1}{\left(c e^{-at}+\frac{1}{(p_{\infty})^{s}}\right)^{\frac{1}{s}}}}
        $$

        Note that if $s = 1$ we are considering the logistic model:
        \begin{gather*}
            \underbrace{\frac{\partial p}{\partial t} = a p(t)\left(1-\frac{p(t)}{p_{\infty}} \right)}_{\text{ODE Richards Model (s=1)}} = a p(t) - \frac{a}{p_{\infty}}     p^{2}(t) \overset{p_{\infty} = \frac{a}{b}}{\Longrightarrow} \\ \overset{p_{\infty} = \frac{a}{b}}{\Longrightarrow} \underbrace{\frac{\partial p}{\partial     t} = ap(t)-bp^{2}(t)}_{\text{ODE Logistic Model}}
        \end{gather*}

        \begin{description}
            \item[Optimized parameters] in view of the above, we consider as the initial values for $a$, $b$ and $c$ those optimized parameters after training the logistic model and $s=1$.

            \item[Implementation] for the optimization of the initial parameters \code{fmin} function from the \code{optimize} package of \code{scipy} library \cite{2020SciPy-NMeth} has been used.
        \end{description}

    \item[Bertalanffy model]
        the Von Bertalanffy growth function (VBGF) was first introduced and developed for fish growth modeling since it uses some physiological assumptions \cite{ramirez1999teoria, de2005fitting}. However, some studies show its possible applications to other types of scenarios, adapting its parameters to be used as a model for population modeling \cite{dawed2014mathematical}. It is therefore reasonable to study the applicability of this model to the evolution of COVID-19 positive cases, as is done in \cite{ahmadi2020modeling}.

        The general formulation of the function is given by the following ODE \cite{fernandes2020parameterizations}:
        \begin{equation}
            \frac{\partial p}{\partial t} = a p^{m}(t) + b p^{n}(t)
        \end{equation}

        Although numerous studies focus only on an appropriate choice of $n$ and $m$ values \cite{renner2018exponent}, as we seek to test the fit of this model, we will take two standard parameters $n=1$ (which is widely assumed \cite{von1957quantitative}) and $m=3/4$ as proposed in \cite{west2001general}. Thus, the explicit solution of the ODE is:
        $$
        \boxed{p(t) = \left(\frac{a}{b}+ce^{\frac{-bt}{4}}\right)^{4}}
        $$

        \begin{description}
            \item[Optimized parameters] $a$, $b$ and $c$ first estimated following a process analogous to that of the Gompertz model.

            \item[Implementation] for the optimization of the initial parameters \code{fmin} function from the \code{optimize} package of \code{scipy} library \cite{2020SciPy-NMeth} has been used.
        \end{description}
\end{description}

\begin{figure}[H]
    \centering
    \vspace*{-0.75cm}
    \includesvg[width=\linewidth]{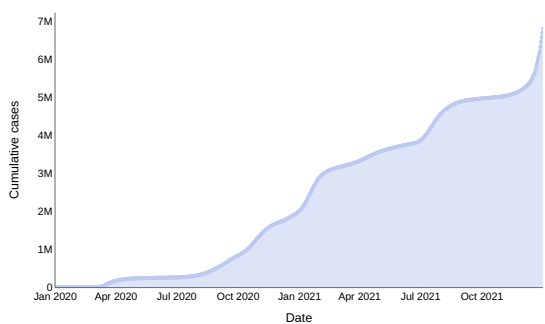}
    \caption{Cumulative COVID-19 confirmed cases in Spain since the start of the pandemic.}
    \label{fig:cum_cases_SP}
\end{figure}

The main motivation for the use of this type of models is the shape of the curve of the cumulative COVID-19 cases. Figure~\ref{fig:cum_cases_SP} shows the cumulative cases in Spain. It can be seen that many sections of the curve follow a sigmoidal shape, which can be modeled, as we have shown, with the previously presented models. Thus, we can take a relatively short period of time (e.g. 30 days), prior to the days we want to predict and apply the previous population models optimizing their parameters to adapt to the shape of the curve and make new predictions. We are especially interested in these new predictions if they are able to capture the trend of the data: periods of ascendance, stability, or descent of the cases.

\subsubsection{Machine learning Models}

After training several machine learning models and testing their predictions on a test set and a validation set, we reduced this battery to the following four widely known models: Random Forest, k-Nearest Neighbours (kNN), Kernel Ridge Regression (KRR) and Gradient Boosting Regressor. All the models under study are minimizing the squared error of the prediction (or similar metrics).

The parameters of each model were optimized using (stratified 5-folds) cross-validated grid-search, implemented with \code{GridSearchCV} from \code{sklearn} \cite{scikit-learn}.

\begin{description}

    \item[Random Forest]
        is an ensemble of individual decision trees, each trained with a different sample (bootstrap aggregation) \cite{ML_book_algorithms}. This type of model is a bagging technique, and the different individual classifiers that it uses (decision trees) are trained without interaction between them, in parallel.
        \begin{description}
            \item[Optimized parameters] the maximum depth of the individual trees, and the number of estimators, i.e., of individual trees in the forest.
            \item[Implementation] \code{RandomForestRegressor} class from \code{sklearn}  \cite{scikit-learn}.
        \end{description}

    \item[k-Nearest Neighbours (kNN)]
        is a supervised learning algorithm, and is an example of instance-based learning. The basic idea of this model is very simple: given a distance (e.g. Euclidean, Manhattan or Hamming distance), the $k$ points of the train set that are closest to the test input $x$ with respect to that distance are searched, to infer what value is assigned to that input \cite{murphy2012machine}.
        \begin{description}
            \item[Optimized parameters] number of neighbors ($k$)
            \item[Implementation] \code{KNeighborsRegressor} class from \code{sklearn} \cite{scikit-learn}.
        \end{description}

    \item[Kernel Ridge Regression (KRR)]
        is a simplified version of Support Vector Regression (SVR). In short, this technique combines Ridge regression (LS and normalization with $l_{2}$ norm), and the kernel trick. For details on this technique, see, e.g., \cite{vovk2013kernel}.

        \begin{description}
             \item[Optimized parameters] $\alpha$ and $\gamma$ (see \cite{krr_sklearn}).
            \item[Implementation] \code{KernelRidge} class from \code{sklearn} \cite{scikit-learn} (with an \code{rbf} kernel).
        \end{description}

    \item[Gradient Boosting Regressor]
        Boosting algorithms combine weak learners into a strong learner iteratively.  Gradient boosting is a boosting-type algorithm for regression \cite{bentejac2021comparative}. It is an ensemble of individual decision trees trained sequentially.
        \begin{description}
            \item[Optimized parameters] learning rate and the number of estimators (i.e. the number of individual trees considered).

            \item[Implementation] \code{XGBRegressor} class from the \code{XGBoost} optimized distributed gradient boosting library \cite{XGBoost}.
    \end{description}

\end{description}

\subsection{Model inputs and outputs}
\label{sec:model_inputs}

The purpose of this work is to generate 14-day forecasts with both population and ML models. In the following sections we will discuss the technicalities of what inputs are needed and how outputs are generated, for each kind of model family.

\subsubsection{Population Models}

Population models are trained with the daily accumulated cases of the 30 days prior to the start date of the prediction. Once fitted with these data, the model outputs the subsequent 14-day prediction (or any amount of desired days for that matter).

Remember that population models use the accumulated cases (instead of raw cases) because it intermittently follows a sigmoidal curve (cf. Figure~\ref{fig:cum_cases_SP}) that these models are especially designed to fit. It should additionally be stressed that population models do not use the rest of the variables (such as mobility, vaccination, etc) that are included in ML models.

\subsubsection{Machine learning Models}
\label{sec:ml-models}

The process of generating predictions with ML models is recurrent. One generates the prediction for the  first day ($n+1$) then one feeds back that prediction back to the model to generate $n+2$, and so on till reaching $n+14$.

To generate a prediction of the cases at $n+1$ the models use previous cases of the last 14 days (\code{lag\textsubscript{1-14}}) as well as the data at $n-14$ for the other variables (mobility, vaccination, temperature, precipitation). We only use $n-14$ and not more recent data ($n$, ..., $n-13$) because these variables have delayed effects on the pandemics evolution.

In the case of the vaccination data, the main motivation for this is that the COVID-19 vaccines manufactured by Pfizer, Moderna and AstraZeneca are considered to protect against the disease two weeks after the second vaccination. With the Janssen vaccine, this value rises to four weeks after the administration of one dose. However, in order to unify criteria (since in this study we will not distinguish by type of vaccine administered), we will consider a two-week delay (see \cite{covid19_vac_Netherlands}).

In order to justify the choice of a lagged effect of 14 days in the case of mobility data, in \cite{MANZIRA2022103770} it is mentioned that scenarios with a lag of two and three weeks of mobility data and COVID-19 infections will be considered for the statistical models. Additionally  \cite{wellenius2021impacts} found that decreases in mobility were said to be associated with substantial reductions in case growth two to four weeks later.

Finally, with respect to the weather data, for example in \cite{chen2020roles} it is mentioned that when considering some meteorological conditions, and four different time delays (1, 3, 7 and 14 days) they authors finally took the weather two weeks ago to model against the daily epidemic situation as its correlation with the outbreak was better. It should be noted that in the latter case we have taken a 7-day rolling average, to reduce the noise and capture the trend in temperature and precipitation (for further details on the climate data pre-processing in Section~\ref{sec:weather}).





The input selection for the recurrent prediction process is illustrated in Table~\ref{tab:input_14days}. Note that the data have been standardized (by removing the mean and scaling to unit variance) using \code{StandandarScaler} from the \code{preprocessing} package of the \code{sklearn} Python library \cite{scikit-learn}.

\begin{table*}[ht!]
    \centering
    \begin{tabular*}{\linewidth}{c @{\extracolsep{\fill}}|ccccc}
          &  \multicolumn{4}{c}{\textbf{INPUT}}\\
         Day  &  &  &  &  & Vaccination/Mobility/ \\
         predicted  & $lag1$ & $lag2$ & ... & $lag14$ & Weather data \\
         \midrule
         $n+1$ & Cases $n$ & Cases $n-1$ & ... & Cases $n-13$ & Data $n-13$\\
         $n+2$ & Pred $n+1$ & Cases $n$ & ... & Cases $n-12$ & Data $n-12$\\
         $n+3$ & Pred $n+2$ & Pred $n+1$ & ... & Cases $n-11$ & Data $n-11$\\
         ... & ... & ... & ... & ... & ...\\
         $n+14$ & Pred $n+13$ & Pred $n+12$ & ... & Cases $n$ & Data $n$\\
    \end{tabular*}
    \caption{Input for predicting data for days $n+1$ to $n+14$ using machine learning models. Note the feedback process taking place in the lags column.}
    \label{tab:input_14days}
\end{table*}

Regarding the variables used as input to the Machine Learning models, different configurations have been tested depending on the data included. Figure~\ref{fig:ml_experiments} (\hyperref[AppendixSupplementaryMaterials]{SupplementaryMaterials}) shows the results obtained with different input configurations. After performing different tests, we decided to analyze the four scenarios exposed in Table~\ref{tab:scenarios}.

\newcommand{\gcheck}{{\color[RGB]{0, 166, 102}\checkmark}}
\begin{table}[H]
    \centering
    \begin{tabular}{c|cccc}
          &  \multicolumn{4}{c}{\textbf{Scenarios}}\\
         \textbf{Input}  & 1 & 2 & 3 & 4  \\
         \midrule
         Cases & \gcheck & \gcheck &  \gcheck & \gcheck  \\
         Vaccination & &  \gcheck & \gcheck & \gcheck \\
         Mobility &   &  & \gcheck & \gcheck \\
         Climate &  &  &   & \gcheck \\
    \end{tabular}
    \caption{Input data for machine learning models. }
    \label{tab:scenarios}
\end{table}

\subsection{Metrics and model ensemble}
\label{sec:model_ensemble}

To evaluate the quality of the prediction we will use the mean absolute percentage error (MAPE) and the root mean squared error (RMSE). The error assigned to a single 14 day forecast is the mean of the errors for each of the 14 time steps.

To make the most of the models' predictions, we use \textit{model ensemble} to aggregate all the individual predictions, thus generating a meta-prediction. The ensemble usually outperforms any of its individual model members \cite{Opitz1999, Rokach2009}. When aggregating predictions we consider the models equally, independently to which family (ML or population) they belong. However, we will show some separate results for each family to highlight the qualitative differences in the predictions they give.

We follow several possible strategies to ensemble the predictions of the models:
\begin{itemize}
    \item \textbf{Mean} prediction of all the models

    \item \textbf{Median} value of the prediction of all models.

    \item \textbf{Weighted average (WAVG)} prediction, where the weight given to each model is the inverse of the RMSE of that particular model on the validation set (cf. Section~\ref{sec:data} for the date ranges of the different splits). That is, the better the performance of a model, the higher the weight we assign to that model.

\end{itemize}

\subsection{Computing environment}
\label{sec:environment}
The computations were performed using the DEEP training platform \cite{8950411}. Also, the study of this work has been implemented using the Python 3 programming language \cite{van1995python}.
In particular, the following additional libraries and versions have been used:
\code{scikit-learn} \cite{scikit-learn} version 0.24.2,
\code{scipy} \cite{2020SciPy-NMeth} version 1.7.1,
\code{pandas} \cite{reback2020pandas} version 1.3.3,
\code{numpy} \cite{van2011numpy} version 1.21.2,
and \code{plotly} \cite{plotly} version 5.3.1.
Additionally \code{flowmap.blue} \cite{flowmapblue} was used to visualize flow maps.

\section{Results and discussion}
\label{sec:results}

\subsection{Results}

We will focus on the results and analysis only in the models trained on Spain as a whole. We nevertheless provide in \ref{appendix_CCAAs} a similar analysis for the 17 Spanish Autonomous Communities.

Tables~\ref{tab:aggregation_scen_SP_MAPE} and \ref{tab:aggregation_scen_SP_RMSE} show the MAPE and RMSE performance for the test set. Columns encode inputs provided to the ML models (cf. Table~\ref{tab:scenarios}) while rows show the different aggregation methods (cf. Section~\ref{sec:model_ensemble}) applied to different subsets of models (\code{ML}, \code{Pop}, \code{All}). Additional plots with model-wise errors are provided in the \hyperref[AppendixSupplementaryMaterials]{Supplementary Materials} (Figure~\ref{fig:modelwise_errors}).


\begin{table}[H]
    \centering
    \resizebox{\linewidth}{!}{
    \begin{tabular}{rl|cccc}
    \toprule
        & & \multicolumn{4}{c}{\textbf{Scenario}}\\
        \multicolumn{2}{c|}{\textit{\textbf{Aggregation}}}   & \textbf{1} & \textbf{2} & \textbf{3} & \textbf{4} \\
        \midrule
         & \textit{ML} & 0.5993 & 0.5571 & 0.5166 & \textbf{0.5052} \\
        \textit{\textbf{Mean}} & \textit{Pop} & 0.5210 & --- & --- & --- \\
         & \textit{All} & \textbf{0.3424} & 0.3501 & 0.3442 & 0.3470 \\
         \midrule
          & \textit{ML} & 0.6224 & 0.5060 & \textbf{0.4610} & 0.4688 \\
        \textit{\textbf{Median}} & \textit{Pop} & 0.5007 & --- & --- & --- \\
         & \textit{All} & \textbf{0.3120} & 0.3352 & 0.3515 & 0.3932 \\
        \midrule
          & \textit{ML} & 0.5831 & 0.5088 & 0.4427 & \textbf{0.4219} \\
        \textit{\textbf{WAVG}} & \textit{Pop} & 0.4954 & --- & --- & --- \\
         & \textit{All} & \textbf{0.3090} & 0.3375 & 0.3363 & 0.3411 \\
    \bottomrule
    \end{tabular}}
    \caption{MAPE obtained in each scenario according to each form of aggregation, for the Spain case in the test split.}
    \label{tab:aggregation_scen_SP_MAPE}
\end{table}

Focusing on the MAPE (Table~\ref{tab:aggregation_scen_SP_MAPE}), one can notice (comparing column-wise) that the Weighted Average aggregation performs better than Median aggregation which in turn performs better than Mean aggregation. Comparing row-wise (\code{ML} rows) we see that adding more variables generally leads to a better performance of ML models. Nevertheless, when we average these ML models with population ones, adding more variables seems to be detrimental (see \code{All} rows).

\begin{figure}[h]
    \centering
    \includesvg[width=\linewidth]{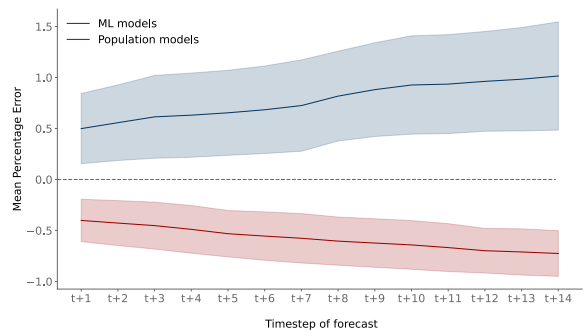}
    \caption{Mean Percentage Error for each time step of the forecast, grouped by model family, for the Spain case in the validation split. Shades show the standard deviation between models of the same family. ML models are trained in Scenario 4.}
    \label{fig:timestep_errors_val}
\end{figure}

The answer to this apparent contradiction comes from looking at the relative error for each model family. For this, in Figure~\ref{fig:timestep_errors_val}, we plot the Mean Percentage Error (i.e. same as MAPE but without taking the absolute value) obtained for each of the 14 timesteps in the validation set. We clearly see that ML models tend to overestimate, while population ones tend to underestimate. This means that when we ensemble both model families, the positive and negative errors cancel out, leading to a better overall prediction. However, this entails that if we improve ML models (by adding more variables as in this case), when we ensemble them with population models the errors end up not cancelling as well. This explains the apparent contradiction that better ML models do not necessarily lead to better ensembles.

It is worth noting than in Figure~\ref{fig:timestep_errors_val}, as expected, both model family errors increase the larger the forecast timestep. But this increase is not evenly distributed, as ML models degrade faster than population models, while their performance is on par at shorter timesteps.

\begin{figure}[h]
    \centering
    \includesvg[width=\linewidth]{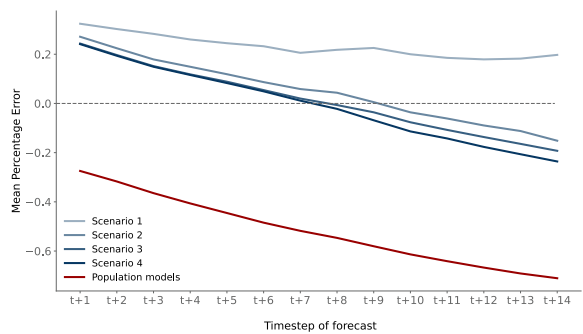}
    \caption{Mean Percentage Error for each time step of the forecast, grouped by model family, for the Spain case in the test split. ML models are shown for the 4 different scenarios.}
    \label{fig:timestep_mpe_test_multi}
\end{figure}

We did the previous analysis on the validation set as it corresponds to a pretty stable phase in COVID spreading, enabling us to clearly identify the over/underestimate behaviour and the performance degradation in both families. The test set however is dominated by an exponential increase in cases due to the sudden appearance of the Omicron variant around mid-November (cf. Figure~\ref{fig:cases_SP_CB}). The patterns still hold, but they are not as straightforward to see. We nevertheless we provide in Figure~\ref{fig:timestep_mpe_test_multi} a similar plot for the test set, showing several ML scenarios. Now, due to the sudden increase in cases, ML models start overestimating, but as the timestep increases, they end up underestimating. On the other side, population models still underestimate, but much more severely than ML models, as expected from the previous analysis. We still see that ML models with more variables tend to individually perform better
\footnote{Scenario 3 seems to perform better than Scenario 4, seemingly contradicting Table~\ref{tab:aggregation_scen_SP_MAPE}, but this is within the expected error of looking at MAPE performances that were already very close in a metric (MPE) that is, obviously, different.}
while the aggregation with population models tends to be worse the more variables you add.
Finally we provide in Figure~\ref{fig:timestep_errors_test}  (in \hyperref[AppendixSupplementaryMaterials]{Supplementary Materials}) a similar plot but subdividing the test set into a stable (\code{no-omicron}) and exponentially increasing (\code{omicron}) phases, where we recover the analysis performed with the validation set.

\begin{table}[h]
    \centering
    \resizebox{\linewidth}{!}{
    \begin{tabular}{rl|cccc}
    \toprule
        & & \multicolumn{4}{c}{\textbf{Scenario}}\\
        \multicolumn{2}{c|}{\textit{\textbf{Aggregation}}}   & \textbf{1} & \textbf{2} & \textbf{3} & \textbf{4} \\
        \midrule
        & \textit{ML} & \textbf{9510.0} & 10121.3 & 10015.4 & 10032.7 \\
        \textit{\textbf{Mean}} & \textit{Pop} & 10006.2 & --- & --- & --- \\
        & \textit{All} & \textbf{9314.0} & 9718.4 & 9700.4 & 9728.0 \\
         \midrule
          & \textit{ML} &  \textbf{9508.8} & 10069.2 & 9935.3 & 9921.0 \\
        \textit{\textbf{Median}} & \textit{Pop} & 9537.2 & --- & --- & --- \\
         & \textit{All} & \textbf{9235.9} & 9693.7 & 9694.1 & 9783.1 \\
        \midrule
        & \textit{ML} & \textbf{9506.9} & 9857.5 & 9602.6 & 9624.1 \\
        \textit{\textbf{WAVG}} & \textit{Pop} & 9713.0 & --- & --- & --- \\
         & \textit{All} & \textbf{9201.0} & 9481.7 & 9427.9 & 9471.4 \\
    \bottomrule
    \end{tabular}}
    \caption{RMSE obtained in each scenario according to each form of aggregation, for the Spain case in the test split.}
    \label{tab:aggregation_scen_SP_RMSE}
\end{table}

For RMSE (Table~\ref{tab:aggregation_scen_SP_RMSE}), comparing column-wise, one still sees that each aggregation method improves on the previous one. But surprisingly, comparing row-wise on \code{ML} rows, we notice that the results go inversely than MAPE results. That is, adding more variables to the ML models leads to worse performance.

Again, this can be explained if we take a closer look at the propagation dynamics during the test split. Recall that, as observed in Figure~\ref{fig:cases_SP_CB}, since mid-November we observe an exponential increase of cases which corresponds to the spread of the Omicron variant.

In Figure~\ref{fig:test_tables_appendix} (\hyperref[AppendixSupplementaryMaterials]{Supplementary Materials}) we split the test results into 2 sub-splits (\code{no-omicron}; \code{omicron}). Then we see that inside each sub-split, RMSE and MAPE follow the same trend, the contradiction disappears. For \code{no-omicron} phase, the best ML scenario is always the one with all the inputs. For \code{omicron} phase, both MAPE and RMSE suggest that the best ML scenario is the one just using cases. This may be due to the importance of the first lags in capturing the significant growth of daily cases. In the full test split, the contradiction appeared because RMSE gives more weight to dates with higher errors (the \code{omicron} phase), while MAPE weights are evenly distributed.

All this analysis seems to suggest that the model is not robust to changes of COVID variant. When it predicts the same variant it was trained on, the model knows how to make good use of all inputs. But when a new variant appears, the spreading dynamics change, and therefore additional inputs just confuse the model, which prefers to rely solely on cases.
Changes in dynamics include facts like Omicron being more contagious (that is same mobility leads to more cases than with the original variant) and being more resistant to vaccines (that is same vaccination levels leads to more cases than with the original variant) \cite{Burki2022}.

\begin{figure}[H]
    \centering
    \includesvg[width=\linewidth]{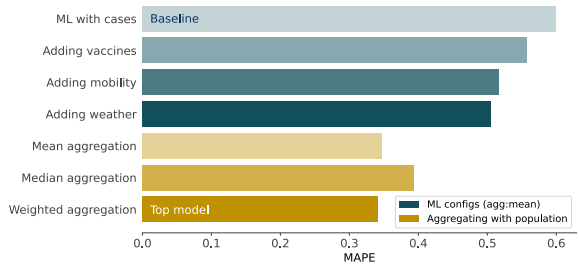}
    \caption{Cumulative improvements for the Spain case in the test split. We color separately (1) improvements made on ML models by adding more inputs (aggregating always with mean), (2) improvements made when aggregating the ML models (with full inputs) with population models with different aggregation methods.}
    \label{fig:ablation}
\end{figure}

Finally, as a visual summary of Table~\ref{tab:aggregation_scen_SP_MAPE} results, we show in Figure~\ref{fig:ablation} how starting with the most basic ensemble (only ML models trained with cases), one can progressively add improvements (more input variables, better aggregation methods), till reaching to the top performing ensemble (ML models trained with all variables and aggregated with population models). The degraded performance with the median aggregation is due to the fact, as discussed earlier, that while ML models improved, the total aggregation with population models happened to be worse.

\subsection{Interpretability of ML models} \label{sec:interpretability}

The interpretability of machine learning models is key in many fields, being the most obvious example the medical or health care field \cite{vellido2020importance}. Understanding the reasons why a model based on artificial intelligence techniques makes a prediction helps us to understand its behavior and reduce its black box character  \cite{rodriguez2020interpretation}. For this purpose, in this work we have used the so-called SHAP values \cite{10.5555/3295222.3295230}. 

SHAP (SHapley Additive exPlanation) values are used to estimate the importance of each feature of the input characteristics space, in the final prediction. The idea is to study the predictions obtained when a feature is removed or added from the model training. Specifically, the final contribution of input feature $i$ is determined as the average of its contributions in all possible permutations of the feature set \cite{rodriguez2020interpretation}. Having a positive/negative SHAP value for input feature $i$ on a given day $t$ means that feature $i$ on day $t$ contributed to pushing up/down the model's prediction on day $t$ (with respect to the expected value of the prediction, computed across the whole training set).

\begin{figure}[H]
    \centering
    \vspace*{-0.75cm}
    \includesvg[width=\linewidth]{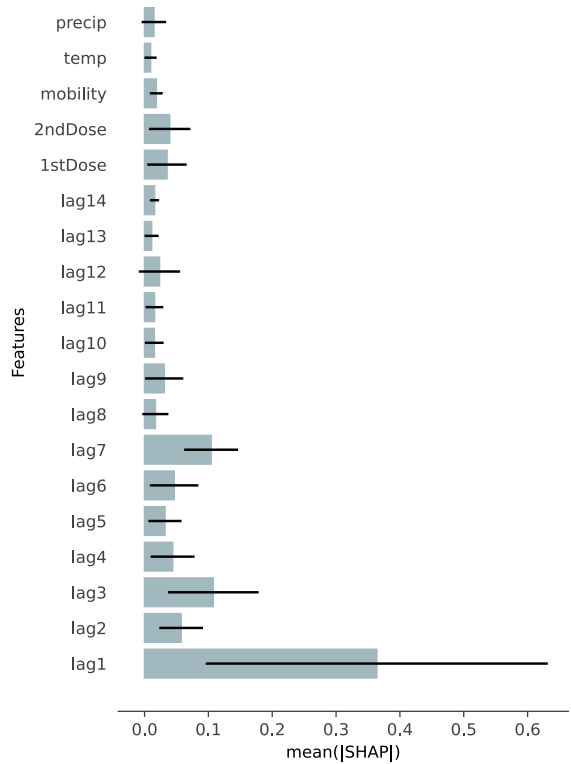}
    \caption{Mean absolute SHAP values (normalized). Error bars show the standard deviation across all the machine learning models.}
    \label{fig:SHAP_values}
\end{figure}

In Figure~\ref{fig:SHAP_values}, we plot the importance of the different features, how much the model relies on a given feature when making the prediction. This importance is computed taking the mean value (across the full dataset) of the absolute value (we do not care whether the prediction is downward or upward) of the SHAP value. This is done feature wise, and we average across the 4 ML models studied (cf. Section~\ref{sec:ml-models}): Random Forest, Gradient Boosting, k-Nearest Neighbors and Kernel Ridge Regression.

We see that the cases lag features, especially the first lags, have the biggest impact on the predictions. As expected, the bigger the lag, the lower the importance of that feature (i.e. more recent the data, the more it matters), with some noisiness in the decrease (e.g. $lag_3$, $lag_7$).

At first glance one might think that non-cases feature (vaccination, mobility and climate), do not matter much compared to the first lags of the cases. This view is obviously biased. The first lags  give a rough estimate of future cases (i.e. future cases are roughly equal to present cases), but the remaining features, while smaller in absolute importance, are crucial to refine the rough estimate upwards or downwards. And this is precisely why we saw that adding more variables always reduced the MAPE of ML models (cf. Table~\ref{tab:aggregation_scen_SP_MAPE})

In Figures~\ref{fig:shap_variablewise_others} and \ref{fig:shap_variablewise_inc} in the \hyperref[AppendixSupplementaryMaterials]{Supplementary Materials}) we provide a more in depth and nuanced overview of the contribution of each feature.

For the case lags, we see that the positive slope in the $lags_{1-7}$ shows that higher lag values correlate with higher predicted cases, which is obviously expected.
For $lags_{8-13}$, this trend is inverted, meaning that higher lag values correlate with lower predicted cases. This is obviously counter-intuitive and we do not have a clear intuition on why this might be happening, but is possibly due to some complex interaction between several features.
In $lag_{14}$ the trend goes back to normal again, suggesting that the model is picking some weekly pattern in the lags (remember that $lag_7$ was also abnormally high) which might be reflecting the moderate weekly pattern we saw in Figure~\ref{fig:boxplots_weekday}.

For non-cases features, we see that:

\begin{itemize}
    \item Mobility does not seem to be strongly correlated with predicted cases. This is possibly due, as stated in Section \ref{subsec:data_limitations}, to the fact that mobility is misleading: when cases grow fast, mobility is restricted, but cases keep growing due to inertia.
    \item Precipitation does not seem to be correlated with predicted cases (probably because precipitation is not a good proxy for humidity).
    \item Higher temperatures are correlated with lower predicted cases as expected (see, for instance, \cite{ROSARIO2020113587}).
    \item Higher number of first vaccine doses are moderately correlated with lower predicted cases as expected, while second doses do not show a mayor correlations. Although unexpected, this lack of negative correlation (more vaccines, lower cases) can be explained by the fact that vaccination efforts tend to increase during peaks in cases, therefore, as with mobility, cases keep growing due to inertia despite vaccination efforts.
\end{itemize}

\subsection{What ended up not working}

Every paper that does not contain its \textit{counterpaper} should be considered incomplete \cite{borges}. Therefore we will devote this section to briefly describe some of the aspects that we have considered, but that ended up not being included in the final model. We also hope to provide, when possible, some insights as for why they did not improve accuracy as expected.

\subsubsection{Input pre-processing}

When deciding the mobility/vaccination/weather lags we where going to use, we tried in each case a number of values based on the lagged-correlation of those features with the number of cases. In the end, the correlation was not an especially good predictor of the optimal lag so we decided to go with the community-standard values (14 lags, cf. Section~\ref{sec:data}).
In addition, we tried to include a weekday variable (either in the $[1, 7]$ range or in binary as weekday/weekend) to give a hint to the model as when to expect a lower weekend forecast. But it did not end up working, possibly due to the fact that the weekly patterns in the number of cases are often buried in the large variations in cases throughout the year (cf. Figure~\ref{fig:boxplots_weekday}).

When we fixed the inputs we were going to use, we tried a number of pre-processing tricks that did not improve the model performance. Among those:
\begin{itemize}
    \item We tried to perform a rolling 7-day average of the mobility to smooth the weekly mobility patterns.
    \item We tried to provide accumulated vaccination instead of raw vaccination. Using cumulative vaccines make more sense than using new vaccines, because one would not expect an sudden increase in cases if for some reason the vaccination stopped for one week, especially if a large portion of the population is already vaccinated.
    \item In addition to the raw features, we tried to add the velocity and acceleration of each feature (cases/mobility/vaccination), to give a hint to the models about the evolution trend of each feature.
\end{itemize}

In the end, all these sensible pre-processing tricks might not have worked because, as we saw in Section~\ref{sec:interpretability}, the correlations between these variables and the predicted cases was not strong enough and their absolute importance was small enough (compared with cases lags) to be distorted by noise.

Finally, regarding the selection of the four scenarios studied, in addition to the configurations discussed above which did not perform successfully, we have tested the seven possible combinations of cases and variables, namely: \textit{\textbf{cases + vaccination}}, \textit{cases + mobility}, \textit{cases + climate}, \textit{\textbf{cases + vaccination + mobility}}, \textit{cases + vaccination + climate}, \textit{cases + mobility + climate} and \textit{\textbf{cases + vaccination + mobility + climate}}. After performing these tests, we decided to take the scenarios shown in Table~\ref{tab:scenarios} because they were the ones that provided the best results.

In Figure~\ref{fig:ml_experiments} (\hyperref[AppendixSupplementaryMaterials]{Supplementary Materials}) we provide a scatter plot with the performance of these additional experiments.

\subsubsection{Output structure}

Regarding the generation of the forecasts, we attempted to generate a single 14-day forecast but it produced substantially worse results. Generating 1-step forecasts and feeding them back to the model, as we finally did, allows the model to better focus and remove redundancies in the predicting task.

\subsubsection{Aggregation methods}

As an additional aggregation method we tried \textit{stacking} \cite{8478522}, where a meta ML model (here, a simple Random Forest) learns the optimal way to aggregate the predictions of the ensemble of models. This meta-model is trained on the validation set (to not favour models that over fit the training set). To be able to have a single meta-model to aggregate both population and ML models, we feed the meta-model with just the predictions of each model for a single time step of the forecast. In other settings, meta-models use both inputs and predictions, but this was not feasible in our case where inputs varied for population and ML models, and across ML scenarios.

In the end, stacking did not improve results, in most cases performing even worse than the simple mean aggregation. This is possibly due to the small size of the validation set, which makes it difficult to learn a meaningful meta-model. Variations of this setup included (1) training a different meta-model for each forecast time step (same performance as single meta-model setup); (2) feeding the meta-model all 14 time steps (worse performance due to noise added by redundant information).

We also tried to a variation of the weighted average in which we weighted models based on their performance on the \code{val} set, but weighting \textit{each time step separately}. This should in principle work better than the standard weighting as it learns to give progressively less weight to models whose forecast degrades more rapidly (that is ML models, cf. Figure~\ref{fig:timestep_errors_val}). In practice it did not show an unequivocal superior performance over the standard weighting, performing in some cases better, in others worse. This is possibly due to the fact that in both setups, weights are computed based on the performance on the \code{val} set, which is relatively small. Therefore one expects, that in the future, with more validation data available, the noise will disappear and this more sophisticated approach will prevail. For the time being, as no clear winner was proclaimed, we decided to favour the simpler approach.

\section{Conclusions}
\label{sec:conclusions}
In this work we have evaluated the performance of four Machine Learning (ML) models (Random Forest, Gradient Boosting, k-Nearest Neighbors and Kernel Ridge Regression), and four population models (Gompertz, Logistic, Richards and Bertalanffy) in order to estimate the near future evolution of the COVID-19 pandemic, using daily cases data, together with vaccination, mobility and climate data. Specifically, our proposal is to use the two families of models (machine learning and population models) to obtain a more robust and accurate prediction.

With regard to the population models, it should be noted that we have used them as an alternative to the compartmental ones because all the data necessary to construct a SEIR-type model were not available for the case of Spain. Despite their simplicity, we have successfully ensemble them with ML models, improving the predictions of any individual model.

We found that, in most cases, when progressively adding more input features, the MAPE error of the aggregation of ML models decreased.
We also saw that this improvement did not necessarily reflected on a better performance when we ensemble them with population models, due to the fact that ML models tended to overestimate while population models tended to underestimate.
When looking at the apparent contradiction in behaviour between MAPE and RMSE metrics, we discovered that ML models did not perform as well on COVID variants different (e.g. Omicron) that the ones they were trained on. When accounting for the change in variant the metrics agreed again.

Finally, we analyzed the SHAP values obtained for each of the 4 ML models to assess the importance of each feature in the final prediction. As expected, this highlighted the importance of recent cases when predicting future cases. Among non-cases features, vaccination and mobility data proved to have significant absolute importance, while lower temperatures showed to be correlated with lower predicted cases. All in all, despite relatively minor absolute importance, non-case features (vaccination, mobility and climate) have proven to be crucial in refining the predictions of ML models.

We foresee several lines to build upon this work.
Firstly, adding more and better variables as inputs to the ML models; for example, introducing data on social restrictions (use of masks, gauging restrictions, ...), on population density, mobility data (type of activity, region's connectivity, etc), or more climatic data such as humidity.
Second, regarding the types of models, we will explore deep learning models, such as Recurrent Neural Networks (to exploit the time-dependent nature of the problem), Transformers (to be able to focus more closely on particular features), Graph Neural Networks (to leverage the network-like spreading dynamics of a pandemic) or Bayesian Neural Networks (to quantify uncertainty in the model's prediction).
All this future work will improve the robustness and explainability of the model ensemble when predicting daily cases (and potentially other variables like Intensive Care Units), both at national and regional levels.

\section*{CRediT author statement}

\textbf{Ignacio Heredia Cacha}: Data Curation, Software, Formal analysis, Visualization, Writing - Original Draft.
\textbf{Judith Sáinz-Pardo Díaz}: Data Curation, Methodology, Software, Formal analysis, Investigation, Visualization, Writing - Original Draft.
\textbf{María Castrillo Melguizo}: Data Curation, Conceptualization, Funding acquisition, Methodology, Software, Writing - Original Draft.
\textbf{Álvaro López García}: Conceptualization, Funding acquisition, Methodology, Software, Supervision, Writing - Original Draft.
\footnote{\url{https://www.elsevier.com/authors/journal-authors/policies-and-ethics/credit-author-statement}}

\section*{Declaration of Competing Interest}

The authors declare that they have no known competing financial interests or personal relationships that could have appeared to influence the work reported in this paper.

\section*{Acknowledgements}

The authors acknowledge the funding and support from
the project Distancia-COVID (CSICCOV19-039) of the CSIC funded by a contribution of AENA;
from the \textit{Universidad de Cantabria} and the \textit{Consejería de Universidades, Igualdad, Cultura y Deporte} of the \textit{Gobierno de Cantabria} via the ``\textit{Instrumentación y ciencia de datos para sondear la naturaleza del universo}'' project;
from the Spanish Ministry of Science, Innovation and Universities through the María de Maeztu programme for Units of Excellence in R\&D (MDM-2017-0765);
and the support from the project DEEP-Hybrid-DataCloud ``Designing and Enabling E-infrastructures for intensive Processing in a Hybrid DataCloud'' that has received funding from the European Union's Horizon 2020 research and innovation programme under grant agreement number 777435.
This research work was also funded by the European Commission - NextGenerationEU (Regulation EU 2020/2094), through CSIC's Global Health Platform (PTI Salud Global).

The authors would also like to thank the Spanish Ministry of Transport, Mobility and Urban Agenda (MITMA) and the Instituto Nacional de Estadística (INE) for releasing as open data the Big Data mobility study and the DataCOVID mobility data. Also, the authors would like to acknowledge the volunteers compiling the per-province dataset of COVID-19 incidence in Spain in the early phases of the pandemic outbreak.

\bibliographystyle{elsarticle-num}
\bibliography{references}

\onecolumn
\newpage
\appendix

\section{Analysis by Autonomous Community}
\label{appendix_CCAAs}

In the following figure the MAPE obtained considering scenario 4 (the most complete) for the machine learning models, and the 3 forms of aggregation: mean, median and weighted mean, is shown for the 17 Spanish autonomous communities.

\begin{figure}[H]
    \centering
    \includesvg[width = 1.12\linewidth]{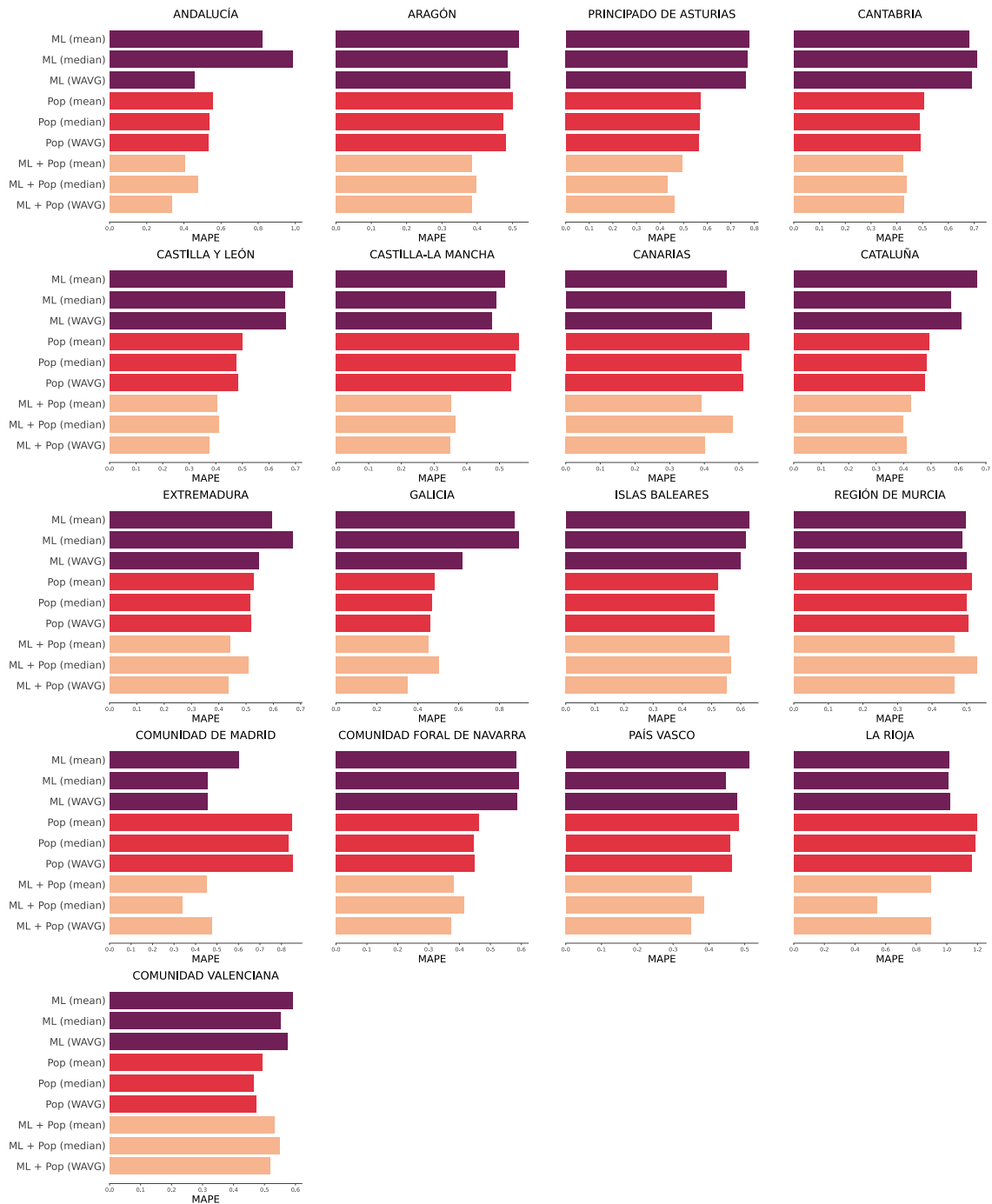}
    \caption{Summary of the MAPE obtained for the cases of the 17 Spanish autonomous communities with the different models. ML models are trained in Scenario 4.}
    \label{fig:MAPE_CCAAs}
\end{figure}

The main aspect to highlight in the previous figure is that in most cases it is still true that the ensemble of models (with any of the three forms of aggregation), manages to improve the MAPE results of the two families of models individually. This particularly clear in Autonomous Communities like Aragón, Cantabria, Castilla y León, Castilla La Mancha, Cataluña or País Vasco.

Regarding whether Machine Learning or population models obtain better results, it is observed that this changes depending on the autonomous community. In some of them, ML models always obtain better MAPE than population ones (e.g. Madrid, Castilla La Mancha, and La Rioja), while in others cases population models obtain better results than ML ones (e.g. Cantabria, Castilla y León and Cataluña).

\section{Supplementary materials}
\label{AppendixSupplementaryMaterials}

\begin{figure}[H]
    \centering
    \includesvg[width=0.9\linewidth]{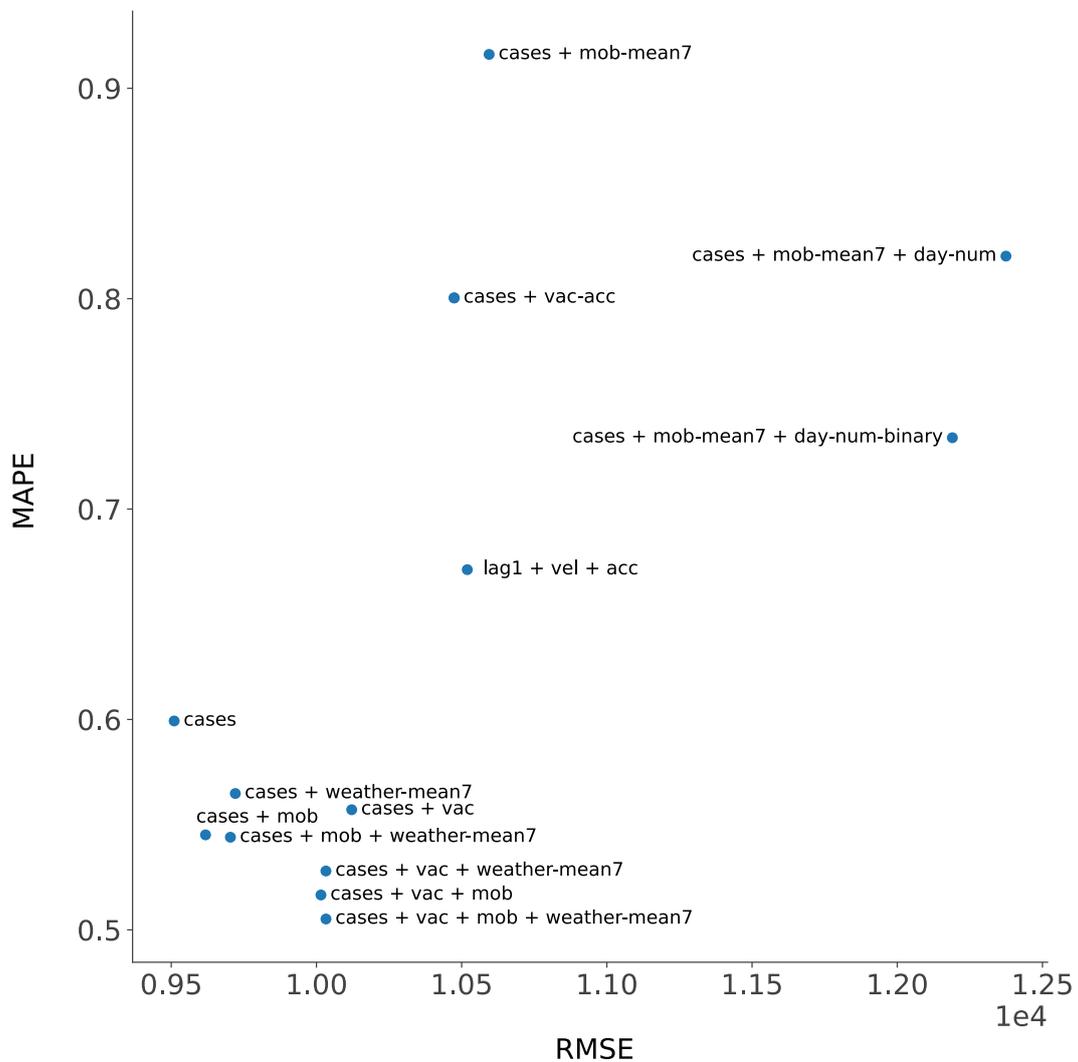}
    \caption{Performance of some of the additional ML configuration we tried, for the Spain case in the test split. Models are aggregated with mean aggregation.}
    \label{fig:ml_experiments}
\end{figure}

\begin{figure}[H]
    \centering
    \includesvg[width=0.6\linewidth]{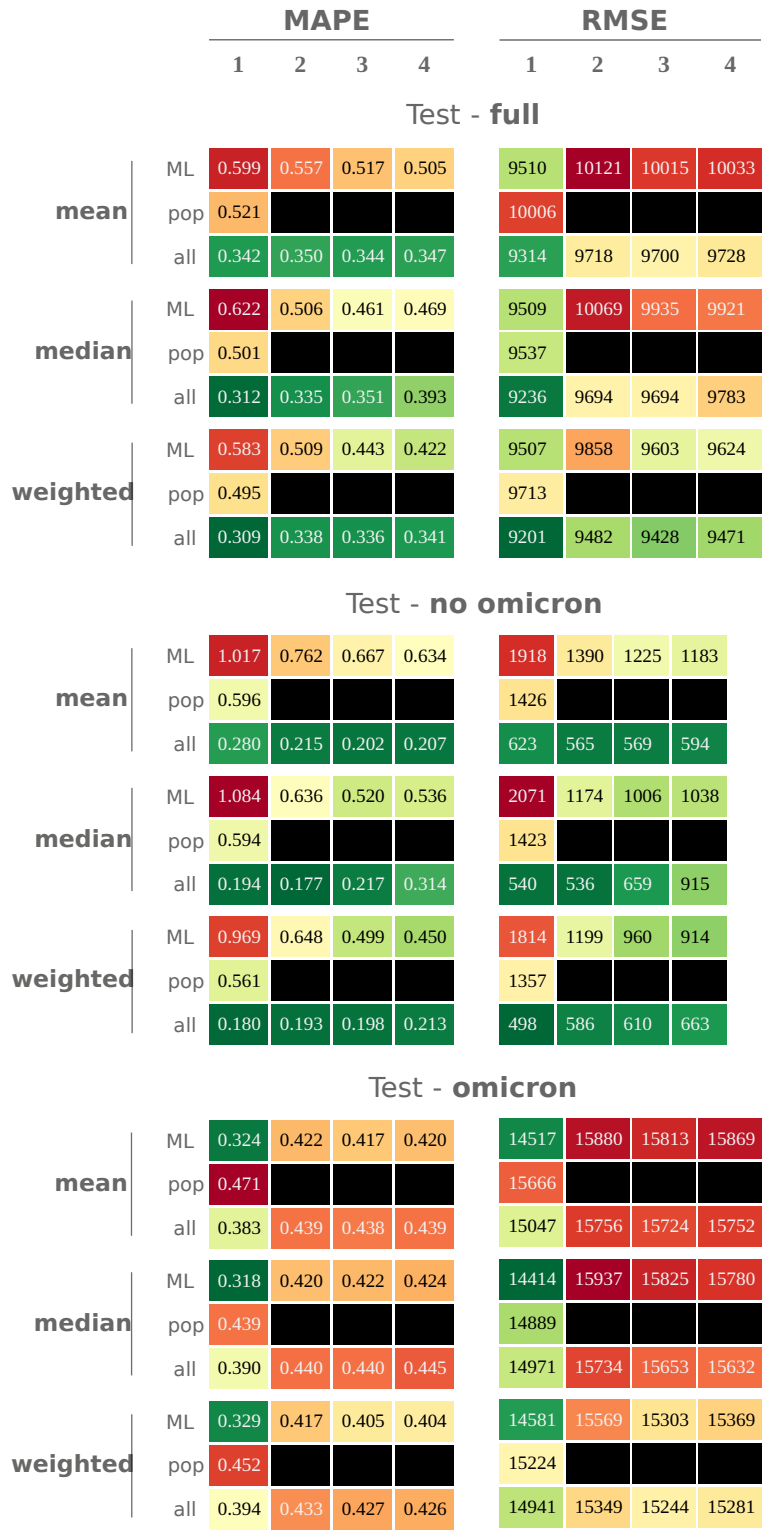}
    \caption{MAPE and RMSE obtained in each scenario (\code{1}, \code{2}, \code{3}, \code{4}) according to each form of aggregation (\code{mean}, \code{median}, \code{weighted}) for different subsets of models (\code{ML}, \code{Pop}, \code{All}).
    We show averages for three different splits: the full test split, the test split \textit{before} Omicron variant started (mid November), and the test split \textit{after} Omicron variant started. All predictions are made for Spain.}
    \label{fig:test_tables_appendix}
\end{figure}

\begin{figure}[H]
    \centering
    \vspace*{-1cm}
    \includesvg[width=0.9\linewidth]{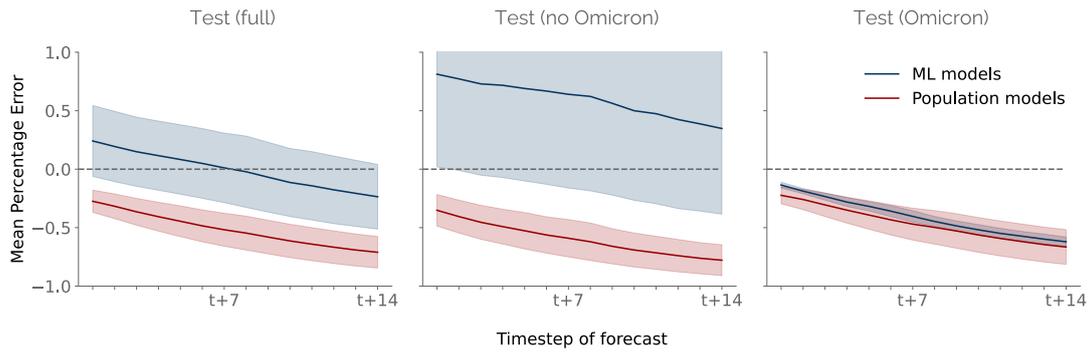}
    \caption{Mean Percentage Error for each time step of the forecast, grouped by model family. Shades show the standard deviation between models of the same family. ML models are trained in Scenario 4. We show results for three different splits: the full test split, the test split \textit{before} Omicron variant started (mid November), and the test split \textit{after} Omicron variant started. All predictions are made for Spain. The \texttt{no-omicron} plot does not exactly reproduce the degradation of ML at large timesteps seen in Figure~\ref{fig:timestep_errors_val} because the split is not as stable as validation (i.e. Omicron is slowly appearing, driving ML models to underestimation).
    }
    \label{fig:timestep_errors_test}
\end{figure}

\begin{figure}[H]
    \centering
    \vspace*{-2cm}
    \includesvg[width=0.9\linewidth]{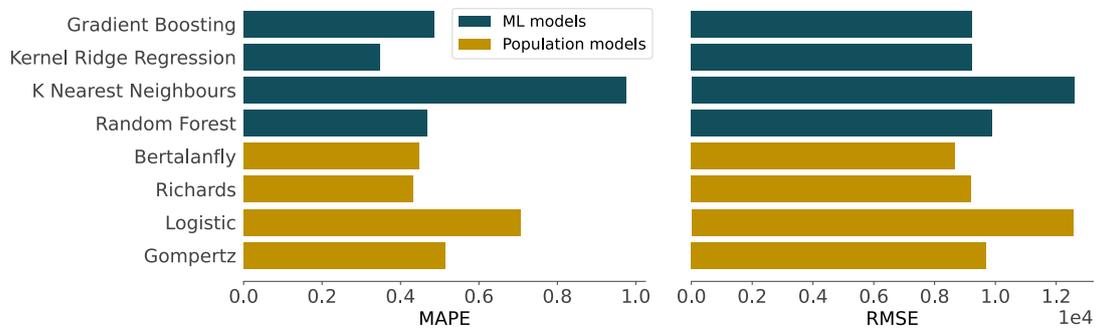}
    \caption{Mean MAPE and RMSE, for each model, for the Spain case in the test split. ML models are trained in Scenario 4.}
    \label{fig:modelwise_errors}
\end{figure}

\begin{figure}[H]
    \centering
    \vspace*{-2cm}
    \includesvg[width=0.9\linewidth]{images/shap_variablewise_others.svg}
    \caption{SHAP dependence plot for non-cases features. For each feature we plot the raw value of the feature vs its associated SHAP value. We average SHAP values across all ML models. All models are trained in Scenario 4 for the Spain case. We display values for all the dataset (train + val + test).}
    \label{fig:shap_variablewise_others}
\end{figure}

\begin{figure}[H]
    \centering
    \includesvg[width=\linewidth]{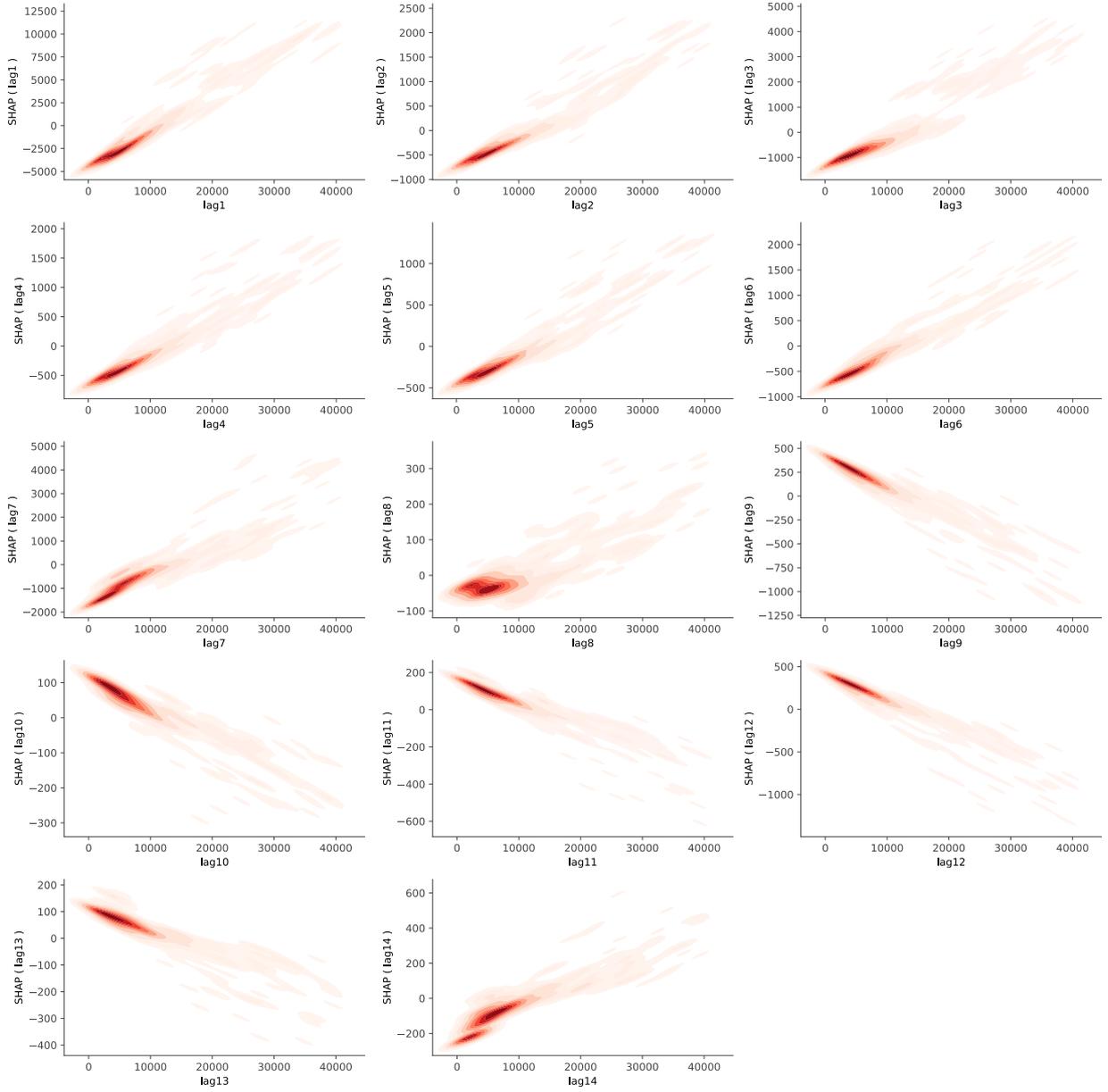}
    \caption{SHAP dependence plot for cases lags features. For each feature we plot the raw value of the feature vs its associated SHAP value. We average SHAP values across all ML models. All models are trained in Scenario 4 for the Spain case. We display values for all the dataset (train + val + test).}
    \label{fig:shap_variablewise_inc}
\end{figure}

\section{Explicit solution of the ODE of the Gompertz model and estimation of the initial parameters}\label{appendix_gomp}

Remember that the ODE which defines the Gompertz model is given by:

\begin{equation}
    \frac{\partial p}{\partial t} = ap(t) -bp(t)log(p(t)),
\end{equation}

being $p(t)$ the population at time t, and $a$ and $b$ two parameters to determine.\\

Taking $y(t)=log(p(t))$, we obtain its explicit solution as follows (note that we consider $y'(t) := \frac{\partial y}{\partial t}$ in order to simplify the notation):

\begin{equation*}
\begin{split}
    \frac{\partial p}{\partial t} = ap(t) -bp(t)log(p(t))  \Longrightarrow & \frac{\partial y}{\partial t} = a -b y(t) \Longrightarrow y'(t) +b y(t) = a \Longrightarrow \\ \Longrightarrow  e^{bt}y'(t)  +e^{bt}b y(t) = e^{bt}a \Longrightarrow & \left(e^{bt}y(t)\right)' = e^{bt}a \Longrightarrow e^{bt}y(t) = a\int e^{bt}  dt \Longrightarrow \\ \Longrightarrow  y(t) = \frac{a}{b}+c e^{-bt} & \Longrightarrow  \boxed{p(t) = e^{\frac{a}{b}+c e^{-bt}}}
\end{split}
\end{equation*}

In order to estimate the parameters $a$, $b$ and $c$ we fix three time instants $t_i$, $t_j$ and $t_k$ verifying: $h = t_{j}-t_{i}$ and $2h = t_{k}-t_{i}$. Be $\displaystyle \alpha = \frac{log(p(t_j))-log(p(t_i))}{log(p(t_k))-log(p(t_i))}$ we get:

\begin{itemize}
    \item $ \displaystyle \alpha = \frac{e^{-b t_{j}} - e^{-b t_{i}}}{e^{-b t_{k}} - e^{-b t_{i}}} \Longrightarrow \alpha = \frac{1}{1 + e^{-bh}} \Longrightarrow   b = -\frac{1}{h}log\left(\frac{1-\alpha}{\alpha}\right)$
    \item $ \displaystyle \frac{p(t_{j})}{p(t_{i})} = e^{c(e^{-bt_{j}}-e^{-bt_{i}})} \Longrightarrow c=\frac{log(p(t_j))-log(p(t_i))}{e^{-bt_j}-e^{-bt_i}}$
    \item $\displaystyle p(t_{i}) = e^{\frac{a}{b} + c^{-bt_{i}}} \Longrightarrow  a=b\left(log(p(t_i))-ce^{-bt_i}\right)$
\end{itemize}

The process to be followed to obtain the initial parameters for the Logistic and Bertalanffy models is analogous to the previous one presented for the Gompertz case.

\end{document}